# Learning Semantic Segmentation with Diverse Supervision

Linwei Ye
University of Manitoba
yel3@cs.umanitoba.ca

Zhi Liu
Shanghai University
liuzhi@staff.shu.edu.cn

Yang Wang
University of Manitoba
ywang@cs.umanitoba.ca

## Abstract

*Models based on deep convolutional neural networks (CNN) have significantly improved the performance of semantic segmentation. However, learning these models requires a large amount of training images with pixel-level labels, which are very costly and time-consuming to collect. In this paper, we propose a method for learning CNN-based semantic segmentation models from images with several types of annotations that are available for various computer vision tasks, including image-level labels for classification, box-level labels for object detection and pixel-level labels for semantic segmentation. The proposed method is flexible and can be used together with any existing CNN-based semantic segmentation networks. Experimental evaluation on the challenging PASCAL VOC 2012 and SIFT-flow benchmarks demonstrate that the proposed method can effectively make use of diverse training data to improve the performance of the learned models.*

## 1. Introduction

Semantic segmentation is one of the key problems in computer vision. The goal is to assign a label to each pixel in the image to indicate the object class of this pixel. Like many areas in visual recognition, the performance of semantic segmentation has improved significantly in the past few years thanks to the adoption of deep convolutional neural network (CNN) [1, 2, 3, 17, 28].

A limitation of current CNN-based semantic segmentation methods is that they require a large amount of training images with pixel-level annotations which are onerous and time-consuming to acquire. This severely restricts the applicability and scalability of existing semantic segmentation methods to real-world scenarios. On the other hand, images with weaker supervision labels (e.g. image-level tags, object bounding boxes) are available at a much larger scale for other computer vision tasks, e.g. image classification [12] and object detection [7]. In addition, these weaker labels are much easier and cheaper to collect than pixel-level labels. For example, collecting bounding boxes for each class in an image is about 15 times faster/cheaper than pixel-level labeling [15]. Unsurprisingly, the image-level label is the easiest to collect, since it only requires annotating whether an object class is present without specifying the exact location in the image.

Several weakly supervised semantic segmentation methods have been proposed to leverage image-level labels [11, 20, 22, 24] or box-level labels [4, 10, 19] for learning semantic segmentation models. These image-level label based models generally employ smooth or constraint terms to infer potential segmentation masks from small and sparse discriminative activation regions which lack dense localization or shape information. On the other hand, box-level label based methods tend to use proposals [23] or unsupervised segmentation approaches [25] to obtain the candidate segmentaion mask from each bounding box. A few related works [8, 13, 27] explore to use both image-level and box-level labels as this paper. Unlike these existing methods, we propose a flexible CNN-based framework to learn semantic segmentation models from diverse data in an end-to-end fashion.

In this paper, we introduce an approach for learning semantic segmentation from diverse data with different types of annotations. In particular, we consider three types of annotations (see Fig. 1) commonly used in computer vision: image-level labels for image classification, box-level labels for object localization/detection, and pixel-level labels for semantic segmentation. Our proposed network consists of three losses to train the network from training images with different types of labels. Available training data with one of these three types of labels can be easily added to our learning scheme and boost the segmentation performance accordingly. In addition, our model is flexible and can be used in combination with any existing CNN-based architectures in semantic segmentation. By exploiting training data with diverse supervision, our method outperforms approaches that only use images with pixel-level supervision.

## 2. Related Work

The performance of semantic segmentation has improved significantly in the past few years due to the success of CNN. Long *et al*. [17] propose to convert fully connected layers into convolutional layers and use in-network upsampling to produce feature map with the desired spatial size, and then make dense labeling prediction of an image based on the feature map. Meanwhile, coarse and fine features are fused between different layers with skip connections to generate fine details for the segmentation results. Chen *et al*. [3] develop a dilated convolution layer to enlarge the field of view (FOV) which is gradually reduced due to the convolution operations. Thus the network can incorporate larger context information without increasing the number of parameters. In this paper, we consider these two segmentation frameworks as our base network.

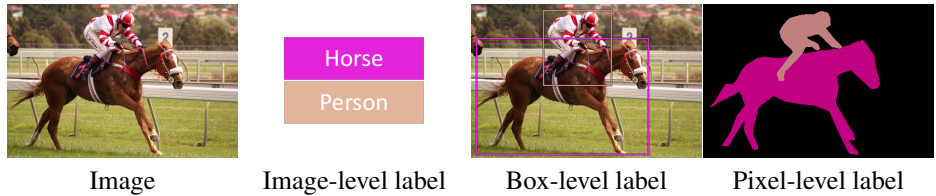

Figure 1. The proposed approach can learn semantic segmentation models by making use of training images annotated for different visual recognition tasks: image-level labels for image classification, box-level labels for object detection, and pixel-level labels for semantic object segmentation.

In order to learn semantic segmentation, we usually need images with pixel-level labels which are very expensive to acquire. To alleviate this limitation, there has been work on weakly supervised semantic segmentation [20, 21, 22, 24]. The goal is to learning semantic segmentation models from labels that are less expensive and easier to obtain, e.g. image-level labels. Some of these approaches [21, 22] are based on multiple instance learning (MIL) framework that considers each image as a bag of pixels and learns semantic segmentation from bag labels (e.g. image-level labels). Pathak *et al*. [20] propose weakly supervised learning with a series of linear constraints. Different linear constraints are incorporated into a loss function on the latent distribution to optimize CNN by standard stochastic gradient descent (SGD). Qi *et al*. [24] introduce an object localization branch to select and aggregate object proposals for suppressing error accumulation in training segmentation network. A new loss that combines seeding, expansion and constrain-to-boundary is proposed in [11]. These three cues are integrated into a segmentation network to exploit image-level labels.

Another line of work on weakly supervised semantic segmentation aims to use box-level labels as the supervision. Dai *et al*. [4] first generate candidate segmentation masks based on object proposal methods, and then iteratively feed these masks as supervision and update masks from the trained network in a recursive training fashion. Papandreou *et al*. [19] develop an EM algorithm with a bias to use weak label images to estimate object labels and iteratively generate latent pixel labels to optimize the CNN parameters. Khoreva *et al*. [10] view the problem as an issue of input label noise and explore training as denoising. Various computer vision processing techniques (including cropping, grabcut segmentation [25] and MCG [23]) are used to obtain effective supervision from box-level labels. Their method can be applied in both semantic segmentation and instance segmentation. Our approach explores a boundary based method to obtain per-pixel labeling of each mask as a soft segmentation and uses the proposed loss function to address ambiguity when a pixel belongs to more than one masks.

In order to exploit both image-level and box-level labels, Kumar *et al*. [13] propose a latent structural support vector machine (LSVM) formulation. They use the latent variables model missing information in the annotation for learning the parameters of a specific-class segmentation model. Guillaumin *et al*. [8] explore the object-background segmentation in the ImageNet [5]. The segmentation process can be propagated at the image and class level by transferring appearance models from the previous stage. Xu *et al*. [27] build a unified model to incorporate different combinations of weak supervision (image level tags, bounding boxes, partial labels) to produce a pixel-wise labeling.

## 3. Our Approach

Given a set of training images with different forms of annotations (image-level labels, box-level labels, pixel-level labels), our goal is to learn a semantic segmentation model from such diverse data. The proposed approach builds on any off-the-shelf semantic segmentation network. Figure 2 illustrates the overall architecture of our approach. The input image of arbitrary size is fed into the fully convolutional network to generate a feature map which has the same size as the input image. Then a separate branch is designed to train the network for each of the three different annotations that the image could have. Note that all the three branches in Fig. 2 operate on the same feature map produced by the same fully convolutional network. So the supervision signals in any of these branches will propagate to the fully convolutional network.

We use FCN [17] as an example segmentation network to present our approach in this section. FCN uses replace the fully connected layers in standard CNN architecture (i.e. VGG [26]) with convolutional layers for semantic segmentation. In our approach, we replace the last softmax layer in FCN with the proposed annotation-specific loss module shown in Fig. 2, so that we can learn the network from diverse data with different annotations. For each input image, the network produces a per-class feature map $f \in \mathbb{R}^{h \times w \times (C+1)}$, where $h$ and $w$ are the height and width of the input image, and $C$ is the total number of object categories (excluding the background class). This feature map can be used to generate the final semantic segmentation, e.g. by assigning the class with the highest value for each pixel. If we have ground-truth per-pixel labels on a training image, we can define a loss function to measure the difference between $f$ and the ground-truth per-pixel labels. However, the challenge is that some of our training images are not annotated at the pixel level, but we still want to utilize all these available data with weaker supervision labels (image-level labels and box-level labels). To this end, we need to somehow convert the feature map $f$ to a different form of prediction label, so that an appropriate loss function can be defined. In our model, the feature map is then fed to one of the three different branches with an annotation-specific loss according to the available label of this input image. In the following, we explain the details of each branch.

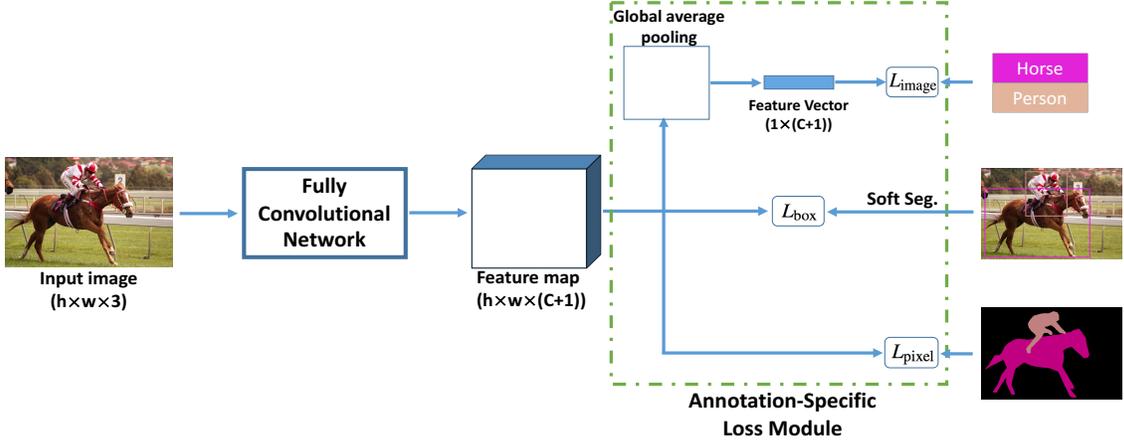

Figure 2. An overview of our approach. Given an image of arbitrary size, a per-class feature map is generated by a fully convolutional network (e.g., FCN [17]). The feature map has the same spatial size as the input image. The number of channels of the feature map corresponds to the number of classes (including the background class). Our model then feeds this per-class feature map to one of three different branches. According to the available label of this input image (e.g. image-level, box-level, or pixel-level), our annotation-specific loss module will use one of the three branches accordingly. In each branch of the annotation-specific loss module, a different loss function is designed according to the form of the label.

### 3.1. Image-level label

This branch (see Fig. 3) is used when the training image only has the image-level label $l = [l_1, l_2, ..., l_C]$, where $l_c$ is a binary label indicating the presence/absence of the object class $c$ in this image. If $l_c = 1$, the object class $c$ is present somewhere in the image. If $l_c = 0$, the object class $c$ is not in the image. Since image-level labels do not contain any localization or shape information of the objects, they can not be used directly to provide supervision for semantic segmentation models which require per-pixel labeling. Instead, we use the feature map $f$ to predict an image-level classification label and define a loss function in terms of the image-level classification. We attach a global average pooling layer [14] on the feature map $f$ to extract a label vector $v \in \mathbb{R}^{1 \times (C+1)}$ in which each element of $v$ corresponds to a specific category, i.e.:

$$v_c = \frac{\sum_{i \in \Omega} f_{i,c}}{|\Omega|} \quad \forall c \in \{0, 1, 2, ..., C\} \quad (1)$$

where $\Omega$ denotes the set of pixels in the image and $f_{i,c}$ is the value of the $i$-th pixel in the $c$-th channel in the feature map $f$.

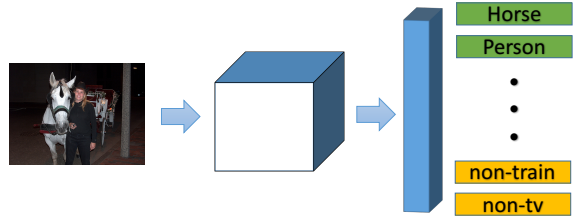

Figure 3. Image-level label training branch. We use the FCN [17] segmentation network and replace the last softmax loss layer with the specific loss module. The length of output vector is the same as the number of classes.

Then we apply a sigmoid activation layer to $v$ and use a multi-category binary cross-entropy loss to measure the

difference between $v$ and the ground-truth label $l$ as follows:

$$L_{\text{image}} = -\frac{1}{C}\sum_{c=1}^{C} l_c \cdot \log(\frac{1}{1+e^{-v_c}}) + (1-l_c) \cdot \log(\frac{e^{-v_c}}{1+e^{-v_c}}) \quad (2)$$

Here, we ignore the loss of the background class which exists in every image so as to push the network to correctly predict object categories.

### 3.2. Box-level label

If an image has the box-level label (i.e. object bounding box), it is sent to the second branch in our annotation-specific loss module. A box-level label provides the rough localization and size of an object in the image, but it does not provide detailed shape information of the object. In this branch, we convert the ground-truth box-level labeling of the training images to a per-pixel labeling in each bounding box. Then we use the per-pixel labeling to define the loss function used in this branch.

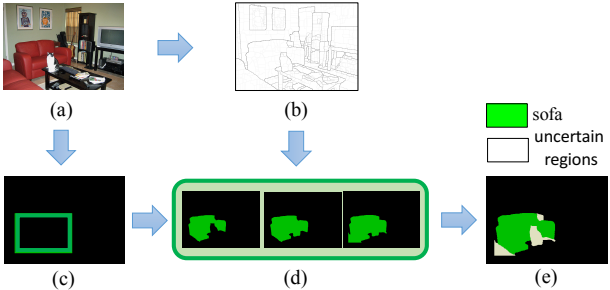

Figure 4. Example of a segmentation mask obtained from the box-level label of an object: (a) input image; (b) UCM; (c) box-level label (i.e. bounding box); (d) multi-scale segmentation; (e) final object mask for this object. This figure is best viewed in color.

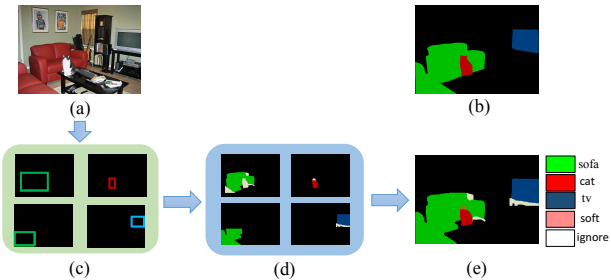

Figure 5. Example of soft segmentation annotation obtained from box-level labels: (a) input image; (b) pixel-level ground truth; (c) box-level labels; (d) object masks corresponding to each of the bounding box; (e) final segmentation mask.

Figure 4 illustrates the process of generating a mask labeling from a box-level label. We leverage ultrametric contour map (UCM) [18] to obtain an object mask of each annotated box. As shown in Fig. 4 (b), UCM assigns different boundary strength values based on region differences.

Darker pixels in Fig. 4 (b) correspond to locations that are more likely to be true object boundaries. Within each annotated box, we first normalize boundary strength to be within $[0, 1]$ and then use three strength thresholds of $\{\frac{1}{4}, \frac{2}{4}, \frac{3}{4}\}$ to generate segmentation masks of different scales (Fig. 4 (d)). From these three coarse-to-fine segmentation masks, we fill regions starting from the inner center region to the maximum occluded boundary and choose a certain mask of a scale until it takes up at least $\alpha\%$ of the region in this box as a confident mask. Finally, we assign the corresponding category label to this confident mask and mark the remaining regions segmented in the coarsest segmentation as uncertain regions. Other regions that do not correspond to foreground in any of these three segmentation masks in the box are considered as background. The generated mask is shown in (Fig. 4 (e)). Our method for generating mask labeling is motivated by the observations in [10]: the area of an object mask should be a reasonable size within an annotated box, and the inner box region should have higher chances of overlapping with the corresponding object. We use $\alpha = 30$ in our experiments.

We also need to consider the ambiguity when a pixel belongs to more than one mask (e.g. in the case of overlapping segmented masks due to overlapping bounding boxes). In order to deal with this case, we define a binary value $s_{i,c}$ to indicate whether the $i$-th pixel belongs to the $c$-th object class (including uncertain regions). If a pixel belongs to two masks with different object classes $c$ and $c'$, we assign $s_{i,c} = 1$ and $s_{i,c'} = 1$ as a soft segmentation pixel. Then we denote $s_i = \sum_{c=0}^{C} s_{i,c}$ and define the loss function for the box-level label as follows:

$$L_{\text{box}} = -\frac{1}{C}\sum_{i\in\Omega}\frac{1}{s_i}\sum_{c=0}^{C}\mathbb{1}(s_{i,c}=1)\log(\frac{e^{f_{i,c}}}{\sum_{j=0}^{C} e^{f_{i,j}}}) \quad (3)$$

The intuition of Eq. 3 is as follows. If a pixel belongs to a single mask, we consider this pixel to be a positive instance of the corresponding object class. But if a pixel belongs to multiple masks of $K$ different object classes, we essentially consider this pixel to be a positive instance of each of the $K$ object classes with a weight of $1/K$. Similar to [10], for the remaining pixels belonging to uncertain regions (i.e. white regions in Fig. 5), we ignore the loss on these pixels during training.

### 3.3. Pixel-level label

If a training image has pixel-level labels, we use the standard fully supervised semantic segmentation learning approach. Let $p_i \in \{0, 1, ..., C\}$ be the label of pixel $i$ in the image. The loss for the pixel-level label image can be calculated by the summation of softmax losses from all pixels

as:

$$L_{\text{pixel}} = -\frac{1}{C} \sum_{c=0}^{C} \sum_{i \in \Omega} \mathbb{1}(p_i = c) \log(\frac{e^{f_{i,c}}}{\sum_{j=0}^{C} e^{f_{i,j}}}) \quad (4)$$

### 3.4. Learning

Let $S_{image}, S_{box}, S_{pixel}$ denote the subset of images with image-level, box-level, and pixel-level labels. Our final loss is the summarization of the losses defined in Sec. 3.1 3.2 3.3 over images in each of the corresponding subset:

$$L_{\text{final}} = \sum_{n \in S_{image}} L_{\text{image}}^{(n)} + \sum_{n \in S_{box}} L_{\text{box}}^{(n)} + \sum_{n \in S_{pixel}} L_{\text{pixel}}^{(n)} \quad (5)$$

Here we use $L_{\text{image}}^{(n)}, L_{\text{box}}^{(n)}, L_{\text{pixel}}^{(n)}$ to denote the corresponding loss defined on the $n$-th training image in $S_{image}$, $S_{box}$ and $S_{pixel}$, respectively.

To optimize the loss function, we use stochastic gradient descent with momentum of 0.9, weight decay of 0.0005 and learning rate of 0.0001. The batch size is set to 1 in order to take an image of arbitrary size as the input and the total iteration is based on the original setting of adopted segmentation framework and adjusted proportionally with the amount of available images. In each mini-batch, we pick images from one of the subsets $S_{image}, S_{box}, S_{pixel}$. In other words, in each mini-batch, only one branch of the annotation-specific loss module is activated.

## 4. Experiments

This section describes the experimental evaluation of our method. We first describe the dataset and the experiment setup in Sec. 4.1. Then we present the experimental results in Sec. 4.2 and comparison with other methods in Sec. 4.3. We perform additional ablation analysis in Sec. 4.4.

### 4.1. Dataset and Setup

**Dataset:** We evaluate the proposed approach on the PASCAL VOC 2012 benchmark dataset [6]. The dataset consists of images of 20 object categories and the background class, and also has instance-level label augmented by [9] to produce box-level annotation. Following the prior work [19, 24], we use the augmented train dataset [9] which contains 10582 images with pixel-level labels. From the original PASCAL VOC 2012 train dataset, we create three subsets (for image-level, box-level, pixel-level). We consider three different ratios of $|S_{\text{pixel}}|, |S_{\text{box}}|, |S_{\text{image}}|$ shown in Table 1.

**Evaluation metrics:** We use the following four evaluation metrics defined in [17] to measure the performance of the semantic segmentation results. Specifically, let $n_{ij}$ be the number of pixels of class $i$ predicted to be class $j$ and $t_i = \sum_j n_{ij}$ be the total number of pixels of class $i$. These four metrics are defined as follows:

- pixel accuracy (pAcc): $\sum_i n_{ii} / \sum_i t_i$
- mean accuracy (mAcc): $(1/C) \sum_i n_{ii}/t_i$
- mean IU (mIU): $(1/C) \sum_i n_{ii}/(t_i + \sum_j n_{ji} - n_{ii})$
- frequency weighted IU (fwIU): $(\sum_k t_k)^{-1} \sum_i t_i n_{ii}/(t_i + \sum_j n_{ji} - n_{ii})$

**Methods for comparison:** We consider FCN [17] and DeepLab [2] as our base networks and reimplement them in pytorch [1]. For each base network, we compare several different variants of our model. Assuming FCN is used as the base network, these variants are defined as follows:

- *FCN-p*: this method ignores the training images with box-level or image-level annotations (i.e. $S_{\text{image}}$ and $S_{\text{box}}$). It only learns the model parameters from the subset $S_{\text{pixel}}$ with pixel-level annotations.
- *FCN-p+i*: this method uses $S_{\text{pixel}}$ and $S_{\text{image}}$, but it ignores $S_{\text{box}}$.
- *FCN-p+b*: this method uses $S_{\text{pixel}}$ and $S_{\text{box}}$, but it ignores $S_{\text{image}}$.
- *FCN-p+b+i* (i.e. our method): this is our method that makes use of all three types of data.

### 4.2. Experimental Results

Table 2 shows the results of learning the FCN model [17] from diverse data on the PASCAL VOC 2012 validation dataset. From the results, we can see that the proposed approach can effectively improve the semantic segmentation performance by exploiting diverse supervision. In particular, using additional weak supervision data leads to better performance. Our approach using all three subsets outperforms other alternatives that only use part of the data. In addition, the relative sizes of the three subsets affect the final performance. If we have more images with pixel-level annotations, the performance is better. This is intuitive since pixel-level annotations provide the most informative supervision for the learning algorithm. When we have more training images with image-level labels and less with pixel-level labels (e.g. when $|S_{\text{pixel}}| : |S_{\text{box}}| : |S_{\text{image}}| = 1 : 5 : 10$), the performance boost of our approach is even more significant.

Our proposed method is also fairly general and can be used with any existing CNN-based semantic segmentation models. We show the generality by replacing FCN [17] with DeepLab [2] (without CRFs). The results are shown in Table 3. Our approach using all available training data outperforms other alternatives that only use part of the training data.

In order to compare the proposed method with fully supervised model, we hereby calculate upper bounds of corresponding variants of our model. The upper bound uses fully pixel-level labels with the same amount of training images to produce the segmentation performance of our model. In other words, this is the best performance we can get if all the training data have pixel-level labels. As shown in Fig. 6, the performance of different variants of our model is comparable to the upper bound performance obtained by learning FCN on the training data using pixel-level labels under different ratios of training data splits. In particular, with the ratio of $|S_{\text{pixel}}| : |S_{\text{box}}| : |S_{\text{image}}| = 1 : 1 : 1$, our model is only about 2% less than the upper bound in terms of mean IU.

---
[1] https://github.com/pytorch/pytorch

| method | annotation | p:b:i=1:1:1 | p:b:i=1:2:3 | p:b:i=1:5:10 |
|---|---|---|---|---|
| FCN-p | pixel | 3527 | 1763 | 662 |
| FCN-p+i | pixel+image | 3527+3528 | 1763+5293 | 662+6610 |
| FCN-p+b | pixel+box | 3527+3527 | 1763+3526 | 662+3310 |
| FCN-p+b+i | pixel+box+image | 3527+3527+3528 | 1763+3526+5293 | 662+3310+6610 |

Table 1. Three different settings of the training data considered in the experiments. Here we use "p:b:i" to denote the ratio of the images in the three subsets $S_{\text{pixel}}$, $S_{\text{box}}$, $S_{\text{image}}$. In each setting, we list the number of training images in each subset.

| Method (p:b:i=1:5:10) | pAcc | mAcc | mIU | fwIU |
|---|---|---|---|---|
| FCN-p | 84.79 | 67.95 | 51.08 | 77.42 |
| FCN-p+i | 85.48 | 68.43 | 53.91 | 77.67 |
| FCN-p+b | 87.18 | 72.25 | 57.07 | 80.20 |
| FCN-p+b+i | **87.47** | **72.65** | **57.51** | **80.80** |
| Method (p:b:i=1:2:3) | pAcc | mAcc | mIU | fwIU |
| FCN-p | 86.73 | 72.41 | 55.58 | 80.13 |
| FCN-p+i | 87.35 | 72.30 | 57.04 | 80.59 |
| FCN-p+b | 88.00 | 74.49 | 58.94 | 81.70 |
| FCN-p+b+i | **88.16** | **75.40** | **58.97** | **82.02** |
| Method (p:b:i=1:1:1) | pAcc | mAcc | mIU | fwIU |
| FCN-p | 88.24 | 72.96 | 58.13 | 81.93 |
| FCN-p+i | 88.52 | 74.20 | 59.73 | 82.26 |
| FCN-p+b | 88.94 | 75.99 | 60.20 | 82.96 |
| FCN-p+b+i | **88.96** | **76.35** | **60.88** | **83.13** |

Table 2. Comparison of different variants of FCN models on the PASCAL VOC 2012 validation dataset.

| Method (p:b:i=1:5:10) | pAcc | mAcc | mIU | fwIU |
|---|---|---|---|---|
| DeepLab-p | 87.64 | 75.50 | 49.56 | 81.80 |
| DeepLab-p+i | 87.99 | 76.17 | 50.08 | 82.68 |
| DeepLab-p+b | 88.31 | 77.56 | 53.44 | 82.82 |
| DeepLab-p+b+i | **88.65** | **78.22** | **53.86** | **83.41** |
| Method (p:b:i=1:2:3) | pAcc | mAcc | mIU | fwIU |
| DeepLab-p | 88.55 | 77.57 | 53.85 | 83.31 |
| DeepLab-p+i | 88.84 | 78.41 | 53.98 | 83.69 |
| DeepLab-p+b | 89.53 | 79.46 | 55.08 | 84.65 |
| DeepLab-p+b+i | **89.60** | **79.70** | **56.43** | **84.74** |
| Method (p:b:i=1:1:1) | pAcc | mAcc | mIU | fwIU |
| DeepLab-p | 89.29 | 78.21 | 56.20 | 84.20 |
| DeepLab-p+i | 89.96 | 80.10 | 56.53 | 85.13 |
| DeepLab-p+b | 90.05 | 80.15 | 57.41 | 85.41 |
| DeepLab-p+b+i | **90.11** | **80.39** | **58.05** | **85.49** |

Table 3. Comparison of different variants of DeepLab models on the PASCAL VOC 2012 validation dataset.

Figure 7 shows some qualitative examples of various FCN models. From the qualitative results, we can see that using the images in $S_{\text{image}}$ (in addition to $S_{\text{pixel}}$) helps to predict exact categories in the images (see examples in row 2 and 5 in Fig. 7). The additional images in $S_{\text{box}}$ help to obtain a better localization and shape for the objects (see examples in the top and bottom rows in Fig. 7). By making full use of all available data with different forms of labels, the proposed approach generally produces more precise semantic segmentation results than using part of them.

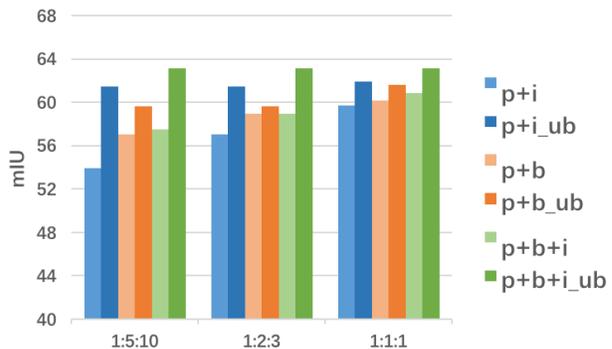

Figure 6. Comparison of different variants of FCN models and their corresponding upper bound on the PASCAL VOC 2012 validation dataset under different ratios of training data splits. $*\_ub$ is the upper bound which uses fully pixel-level labels with the same amount of training images to learn the FCN model. The lighter colors show the results of different variants of the proposed model and the corresponding darker colors show upper bound results.

### 4.3. Comparison with Other Methods

In Table 4, we compare our model with [27] which is also developed for semantic segmentation under various forms of supervision. Since [27] uses the SIFT-flow dataset [16] in the experiment, we apply our approach on the SIFT-flow in order to compare with [27]. SIFT-flow contains 2,688 images with pixel-level labels for 33 semantic categories, including 2,488 for training and 200 for testing. Following the setting in [27], we use a sampling ratio of 0.5 for image-level labels and pixel-level labels respectively. Our model outperforms [27] by a large margin in terms of pixel accuracy and mean accuracy.

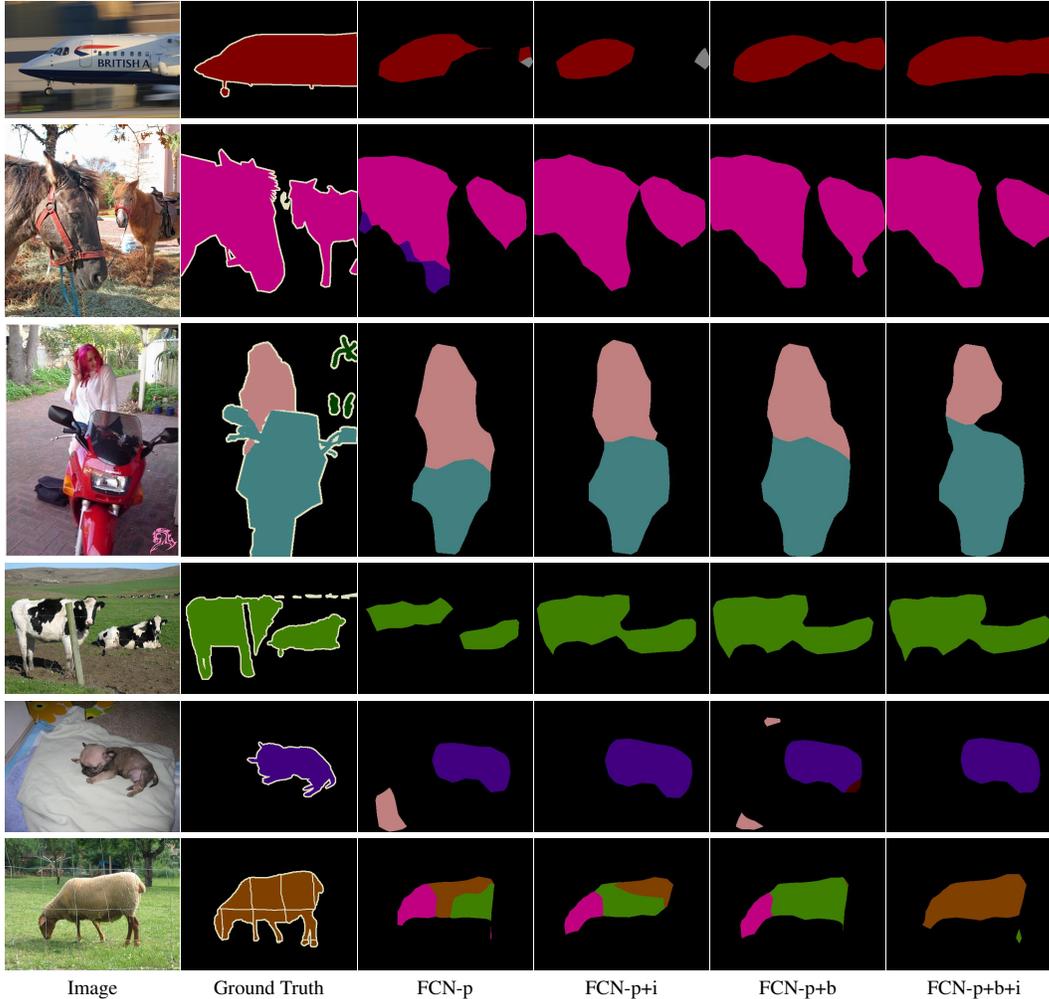

| Image | Ground Truth | FCN-p | FCN-p+i | FCN-p+b | FCN-p+b+i |

Figure 7. Some qualitative examples of our approach on the PASCAL VOC 2012 validation dataset. In each row, from left to right: Image, Ground Truth and segmentation results generated by FCN-p, FCN-p+i, FCN-p+b and FCN-p+b+i.

We also compare our results with the state-of-the-art weakly supervised semantic segmentation methods [4, 10, 19] that use box-level labels. Similarly, we follow their settings with the same numbers of pixel-level and box-level labels in training on PASCAL VOC 2012. Table 5 shows the semantic segmentation results. Note that all three compared methods are originally proposed for weakly supervised semantic segmentation rather than diverse supervision. Since the source codes of these methods are not available, we directly take the performance numbers in those papers. However, these methods in Table 5 use different backbone networks and some of them apply CRF post-processing to refine the results. So it is difficult to have a fair comparison directly. This can be seen from the fact that the fully supervised versions of these methods give slightly different performance numbers (see Table 5). Although our model is not specifically designed for weakly supervised semantic segmentation, it can still achieve comparable performance in this setting.

| Method | pAcc | mAcc |
|---|---|---|
| Xu *et al*. [27] | 72.8 | 36.5 |
| Ours | **82.08** | **58.76** |

Table 4. Comparison to [27] on the SIFT-flow dataset under the same setting (p:i=1:1).

### 4.4. Ablation Analysis

In this section, we perform additional ablation analysis of our approach. We conduct experiments on the FCN model under the ratio of $|S_{\text{pixel}}| : |S_{\text{box}}| : |S_{\text{image}}| = 1 : 5 : 10$, which has the least amount of pixel-level labels and the method has to rely more on images with the box-level or

| Method | val. semi / full | test. semi / full |
|---|---|---|
| WSSL [19] | 65.1 / 67.6 (96.30) | 66.6 / 70.3 (94.74) |
| BoxSup [4] | 63.5 / 63.8 (99.53) | 66.2 / - (-) |
| SDI [10] | 65.8 / 69.1 (95.22) | 66.9 / 70.5 (94.89) |
| Ours | 60.14 / 63.14 (95.25) | 61.42 / 62.2 (98.75) |

Table 5. Comparison of different weakly-supervised segmentation methods on the PASCAL VOC 2012 validation and test datasets under the same setting as [10] in terms of mIU. The numbers of pixel-level and box-level annotations are 1.4k and 9k, respectively. For each method, we show the mIU in both semi-supervised and fully-supervised settings. Since different methods use different design choices (such as backbone network, CRF postprocessing, etc), it is difficult to compare their performance numbers directly. This can be seen from the fact that these different methods have different mIU in the fully supervised setting. The numbers in (·) indicate the ratio of the mIU between the semi-supervised and fully-supervised settings. In the ideal case, this number will be close to 100 indicating that the semi-supervised setting achieves the upper-bound (full-supervised) performance.

| FCN | | pAcc | mAcc | mIU | fwIU |
|---|---|---|---|---|---|
| Pseudo pixel-level | Grabcut | 86.68 | 71.91 | 55.60 | 79.78 |
| | MCG | 86.71 | 70.87 | 56.19 | 79.51 |
| | Hard seg. | 87.30 | 71.80 | 57.35 | 80.33 |
| Our loss | Raw box | 87.04 | 71.07 | 56.91 | 79.87 |
| | Ours | **87.47** | **72.65** | **57.51** | **80.80** |

Table 6. Comparison of different strategies to use box-level labels in FCN on the PASCAL VOC 2012 validation dataset.

image-level labels.

First, we explore and compare other alternative ways of exploiting the box-level supervision. We consider four different ways of using box-level supervision as follows:

**Hard segmentation:** this method uses the same approach to obtain the mask labeling from box-level labels as Sec. 3.2, but it does not keep all overlapping segmented masks of different object classes and weight them in the loss. Instead, every pixel has to belong to a certain class. We randomly assign a class to the overlapping region of multiple segmented masks and then uses it as the pseudo pixel-level label.

**GrabCut:** this method takes each box-level label and applies GrabCut [25] to obtain the binary mask of the object inside the bounding box. It then considers the binary mask as the pseudo pixel-level labels.

**MCG:** this method uses the MCG algorithm [23] to generate a set of candidate object masks for an input image. Inside each box-level label, an object mask that has the maximum overlap with the corresponding bounding box is selected as the object mask for the object inside this bounding box. Similarly, this object mask is used as the pseudo pixel-level labels. This strategy has been used in [10].

**Raw box:** this method assigns the class label to each pixel within a bounding box and considers the pixels outside of any bounding boxes in the image as background. Therefore, Eq. 3 can also be adopted to address ambiguity caused by overlapping bounding boxes.

Table 6 shows the results of different strategies of using box-level labels in the proposed approach, respectively. The segmentation performance with our proposed box-level supervision outperforms all other alternatives. In addition, the proposed box-level supervision loss in Eq. 3 shows its effectiveness to address the ambiguity caused by overlapping regions with multiple uncertain labels. Even using raw box method in the box-level supervision, the semantic segmentation performance is comparable to their pseudo pixel-level methods which require additional pre-processing with segmentation or proposal.

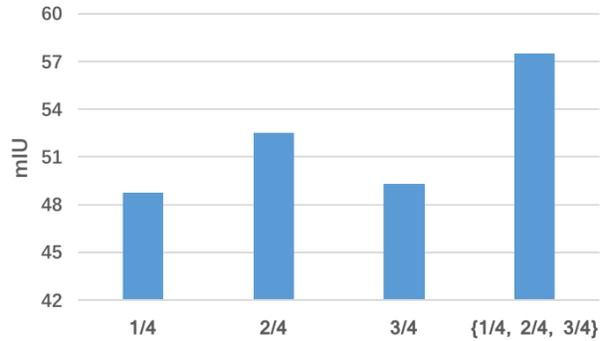

Figure 8. Effect of using three strength thresholds of scales to generate segmentation masks.

We also analyze the effects of some of the hyperparameters in our method. Figure 8 shows the performance of using one strengh threshold ($\frac{1}{2}$, $\frac{2}{4}$, or $\frac{3}{4}$) versus using all three strength thresholds (our method) when generating the segmentation masks. We can see that using all three strength thresholds outperforms just using one threshold.

## 5. Conclusion

We have proposed a method for learning CNN-based semantic segmentation models from training images with diverse annotations, including image-level labels, box-level labels and pixel-level labels. The proposed model consists of three specific losses for each of three annotations and can be easily used together with any existing CNN-based semantic segmentation networks. The experimental results show that our method can effectively make use of diverse training data with different levels of supervision.

**Acknowledgement:** This work was supported by NSERC, the National Natural Science Foundation of China under Grant No. 61771301, and the University of Manitoba GETS funding program. We also thank NVIDIA for the GPU donations.